%% file: egpaper_for_review.tex
\PassOptionsToPackage{prologue,dvipsnames}{xcolor}
\documentclass[10pt,twocolumn,letterpaper]{article}

\usepackage{iccv}
\usepackage{times}
\usepackage{epsfig}
\usepackage{graphicx}
\usepackage{subcaption}
\usepackage{amsmath}
\usepackage{amssymb}

\usepackage{booktabs}
\usepackage{multirow}
\usepackage{xcolor}
\usepackage{bm}
\usepackage{multirow,multicol, makecell, booktabs}
\usepackage{wrapfig}

\usepackage[accsupp]{axessibility}  

\usepackage[dvipsnames]{xcolor}
\usepackage{pifont} 

\usepackage{tikz}
\usepackage{enumitem}
\usepackage[T1]{fontenc}
\usepackage{mathabx}

\newcommand\Tstrut{\rule{0pt}{2.3ex}}
\input{section/math_commands.tex}

\usepackage[pagebackref=true,breaklinks=true,letterpaper=true,colorlinks,bookmarks=false]{hyperref}

\iccvfinalcopy 

\def\iccvPaperID{5388} 

\ificcvfinal\fi

\begin{document}

\title{VLN-PETL: Parameter-Efficient Transfer Learning for \\Vision-and-Language Navigation}

\author{
Yanyuan Qiao \quad  Zheng Yu  \quad Qi Wu\thanks{Corresponding author}\\
Australian Institute for Machine Learning, The University of Adelaide\\
{\tt\small \{yanyuan.qiao,zheng.yu,qi.wu01\}@adelaide.edu.au}\\
}

\maketitle
\ificcvfinal\thispagestyle{empty}\fi

\begin{abstract}
The performance of the Vision-and-Language Navigation~(VLN) tasks has witnessed rapid progress recently thanks to the use of large pre-trained vision-and-language models. However, full fine-tuning the pre-trained model for every downstream VLN task is becoming costly due to the considerable model size. Recent research hotspot of Parameter-Efficient Transfer Learning (PETL) shows great potential in efficiently tuning large pre-trained models for the common CV and NLP tasks, which exploits the most of the representation knowledge implied in the pre-trained model while only tunes a minimal set of parameters. However, simply utilizing existing PETL methods for the more challenging VLN tasks may bring non-trivial degeneration to the performance. Therefore, we present the first study to explore PETL methods for VLN tasks and propose a VLN-specific PETL method named VLN-PETL. Specifically, we design two PETL modules: Historical Interaction Booster (HIB) and Cross-modal Interaction Booster (CIB). Then we combine these two modules with several existing PETL methods as the integrated VLN-PETL. Extensive experimental results on four mainstream VLN tasks (R2R, REVERIE, NDH, RxR) demonstrate the effectiveness of our proposed VLN-PETL, where VLN-PETL achieves comparable or even better performance to full fine-tuning and outperforms other PETL methods with promising margins. The source code is available at \href{https://github.com/YanyuanQiao/VLN-PETL}{https://github.com/YanyuanQiao/VLN-PETL}
\end{abstract}

\input{section/introduction}
\input{section/related_work}

\input{section/method}
\input{section/experiments}
\input{section/conclusion}

{\small
\bibliographystyle{ieee_fullname}
\bibliography{egbib}
}

\end{document}

%% file: section/math_commands.tex
\usepackage{amsmath,amsfonts,bm}

\def\eqref#1{equation~\ref{#1}}

\def\1{\bm{1}}

\def\vf{{\bm{f}}}

\def\vh{{\bm{h}}}

\def\vo{{\bm{o}}}

\def\mQ{{\bm{Q}}}

\def\mV{{\bm{V}}}
\def\mW{{\bm{W}}}
\def\mX{{\bm{X}}}

\DeclareMathAlphabet{\mathsfit}{\encodingdefault}{\sfdefault}{m}{sl}
\SetMathAlphabet{\mathsfit}{bold}{\encodingdefault}{\sfdefault}{bx}{n}

\def\sR{{\mathbb{R}}}



%% file: section/introduction.tex
\section{Introduction}
\label{sec:intro}

Large-scale pre-trained models have shown remarkable success in both computer vision (CV) and natural language processing (NLP) domains, and have largely improved the performance of a variety of visio-linguistic tasks~\cite{itr,NiuZLC21REC,reg}.
These models follow a standard pretrain-and-finetune paradigm, which first pretrains the model on large-scale unlabeled data and then finetunes it on each downstream task. Since the size of such models is growing  rapidly~\cite{gpt,clip}, even fully finetuning and storing a copy of the entire pretrained model for each downstream becomes costly.

\begin{figure}[!t]
	\begin{center}
		\includegraphics[width=1.0\linewidth]{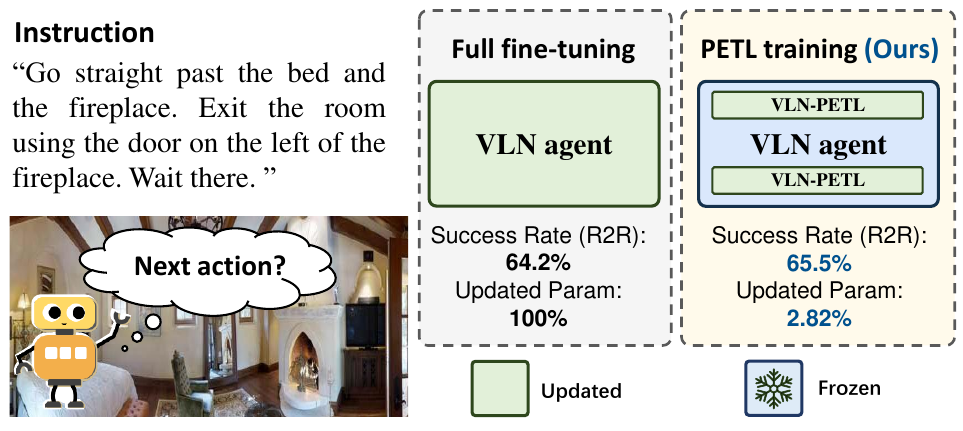}
	\end{center}
	\vspace{-20pt}
	\caption{Comparison of full fine-tuning and our proposed PETL training for VLN tasks. By updating only a small subset of parameters, our proposed VLN-PETL can achieve a comparative performance compared to full fine-tuning} 

	\label{fig:pic1}
	\vspace{-19pt}
\end{figure}

To alleviate this problem, Parameter-Efficient Transfer Learning (PETL) has been proposed as an alternative training strategy~\cite{BenZaken2022BitFitSP,HeLYTDCLBS20,houlsby2019adapter, hu2021lora,Mahabadi2021CompacterEL,MahabadiR0H20} and initially achieved great progress in NLP community. These methods aim to exploit the representation knowledge in the large pretrained models by freezing most parameters of the model and only tuning a small set of parameters, which can achieve comparable or even better performance to full fine-tuning. 
Several approaches have attempted to apply PETL techniques to CV and V\&L domains~\cite{KroneckerAdaptation,st-adapter,lst,sung2022vladapter} and achieved promising results on various downstream tasks. %
Recent works \cite{unified-petl,unipelt,zhang2022hyperpelt} find that different PETL methods have different characteristics and performance on the same downstream task and thus combining multiple PETL techniques may be more effective in improving the performance.

Vision-and-Language Navigation (VLN), which deals with visual, linguistic and robotic action inputs simultaneously, could benefit from the pre-trained large models while suffering from the considerable model size during the downstream tasks fine-tuning. Considering downstream VLN agents are complex enough, full finetuning them with the large pre-trained models for each downstream VLN task becomes expensively, in which case the technique of PETL shows great potential.
Unlike most NLP, CV, and V\&L tasks, VLN is a dynamic action decision-making task relying on the current environment and previous history knowledge of the chosen actions. Specifically, given the instruction in natural language, the VLN agent perceives a new visual observation according to the chosen action at the previous timestep and should choose the next action to perform at the current timestep. Thus, how to effectively learn history knowledge is crucial to adapting PETL methods for VLN tasks. Moreover, the cross-modal interaction which plays a vital role in action prediction should be also enhanced during the process of efficient tuning. In addition, our experiments show that directly applying some existing PETL methods to VLN tasks may bring non-trivial performance degeneration.

Considering these reasons, we propose a VLN-specific PETL method named VLN-PETL. 
Specifically, we design two tailored PETL modules for VLN: Historical Interaction Booster (HIB) and Cross-modal Interaction Booster (CIB). Both these two modules mainly consist of bottleneck layers and multi-head cross-attention layers.
HIB enhances the interaction between the observation and the previous historical knowledge in a recurrent pattern.
While CIB adopts a two-stream structure to focus on the interaction of cross-modal knowledge. 
Similar to adapters which inject bottleneck layers into transformer blocks for efficient tuning, we insert HIB and CIB into the visual encoder and cross-modal encoder separately in the pre-trained model for VLN. During the training process, the original weights of the large pre-trained model are frozen and only weights of these newly injected modules are trained and updated for different downstream VLN tasks.
In addition to HIB and CIB, VLN-PETL also adopts vanilla adapters to efficiently tune the language encoder and the LoRA to further improve the parameter-efficient tuning's performance as previous work declared~\cite{unified-petl,unipelt,zhang2022hyperpelt} for downstream VLN tasks.

We conduct extensive experiments on four mainstream VLN tasks: R2R~\cite{r2r}, REVERIE~\cite{reverie}, NDH~\cite{ndh}, and RxR~\cite{rxr}. The results show that VLN-PETL not only surpasses other PETL methods with promising margins but also achieves comparable or even better performance compared to full fine-tuning, especially on R2R ($\uparrow1.3\% $ SR on validation unseen set, updating only 2.82\% params, see Figure~\ref{fig:pic1}) and on NDH ($\uparrow1.08$ GP on test unseen set which achieves the top position in the leaderboard). We also conduct ablation studies to evaluate the contribution of each component of VLN-PETL and validate the superiority of HIB and CIB to counterpart PETL methods.

In summary, our contributions are as follows: (1) We present the first study that explores Parameter-Efficient Transfer Learning (PETL) techniques for Vision-and-Language Navigation (VLN) tasks; 
(2) We propose a VLN-specific PETL method named VLN-PETL, which incorporates existing PETL methods with two tailored PETL modules for VLN tasks: Historical Interaction Booster (HIB) and Cross-modal Interaction Booster (CIB); (3) Extensive experiments on four VLN downstream tasks demonstrate the effectiveness of our proposed VLN-PETL, which outperforms other PETL methods and keep competitive to full fine-tuning with much fewer trainable parameters.

%% file: section/related_work.tex
\section{Related work}
\label{sec:related_work}

\vspace{-3pt}
\noindent\textbf{Vision-and-Language Navigation}
In the past few years, VLN has received great attention and many methods have been proposed~\cite{Hong_2022_CVPR,recurrent,Lin2022ADAPTVN,Lin2021MultimodalTW,orist,oaam}. Early works were mainly based on the encoder-decoder frameworks~\cite{r2r,speakerfollower,selfmonitor}. While subsequent works approached VLN research in a variety of ways, such as data augmentation~\cite{speakerfollower,Li2022EnveditEE,rem} to improve the robustness of the agent, progress monitoring~\cite{selfmonitor,softexp,Zhu0CL20} to estimate the completeness of instruction-following, and back-tracking~\cite{ke2019tactical,regretful} to help the agent learn to decide when to perform backtracking depending on the state of the agent. Recently, BERT-based pre-training methods have significantly improved the agents' performance in VLN tasks~\cite{airbert,press}. These methods follow the pretraining-and-finetuning paradigm, which first pre-trains a vision-and-language model on a great many text-image pairs of instructions and trajectories with specifically designed proxy tasks, and then fully finetunes the pre-trained model for every downstream VLN task.
VLNBERT~\cite{bertvln} first utilizes this paradigm to solve different downstream VLN tasks, which introduces an extra proxy task of scoring path-instruction pairs in addition to the Masked Language Modeling task. 
PREVALENT~\cite{prevalent} introduces a single-step action prediction proxy task, aiming to learn action-oriented generic visio-linguistic representation. 
HOP~\cite{hop} and HOP+~\cite{qiao2023hop+} exploit past observations to enhance the learning of temporal order modeling and historical information by introducing three VLN-specific proxy tasks. 
HAMT~\cite{hamt} encodes past panoramic observations as historical information explicitly. 

Though these pretraining-and-finetuning methods have attempted to utilize vital historical knowledge for action prediction, the exploration is still limited due to the gap between pretraining and finetuning. Meanwhile, these methods face the same challenge in the fine-tuning stage: oversized parameters for efficient tuning. Specifically, all parameters of the full pre-trained model will be trained and stored for each downstream VLN task. This may hinder the application of VLN in real-world scenarios since realistic robots will have difficulty in training and storing such huge parameters for every new task. Recent works of Parameter-Efficient Transfer Learning (PETL) show great potential in solving this problem, which freeze most parameters in the large pre-trained model while only training and storing minimal parameters for every downstream task.
Thus, we propose the first work that studies and applies PETL methods to VLN named VLN-PETL.

\noindent\textbf{Parameter-Efficient Transfer Learning}
Recently, with the rapid increase of pre-trained models' size, how to efficiently tune the large pre-trained models has received great attention and Parameter-Efficient Transfer Learning (PETL) has become a popular research area. 
One category of PETL methods adds new parameters into the pretrained model and only trains these parameters. For example, Adapter~\cite{houlsby2019adapter} introduces bottleneck layers after attention layers and feed-forward layers in the transformer block. LoRA~\cite{hu2021lora} injects trainable low-rank decomposition matrices into linear projection layers to approximate the update of large amount of parameters. Prompt Tuning~\cite{lester-etal-2021-power} prepends trainable prompts to the model's input. Another kind of PETL method does not add new parameters and selects a subset of the pretrained model's parameters to update, such as BitFit~\cite{BenZaken2022BitFitSP}, which only trains the bias term in the model. Recent works \cite{unified-petl,unipelt,zhang2022hyperpelt} find that incorporating different PETL methods as sub-modules may help improve the integrated performance for different downstream tasks in NLP and CV.

However, VLN is a dynamic task of action prediction relying on both the current observations and previous action decisions, which is more challenging than other static NLP, CV and V\&L tasks. In other words, the previous decision of action will influence the current environment observed by the agent as well as the choice of the next action to perform. Thus, it is important to effectively utilize the historical knowledge of the previous trajectory when applying PETL methods to VLN. Furthermore, the cross-modal interactions of the language and vision also have a great impact on action predictions, which should be enhanced especially when most parameters of the pre-trained model are frozen for efficient tuning. Therefore, we specifically design two PETL modules for VLN tasks to enhance the history knowledge interactions and cross-modal knowledge interactions, namely Historical Interaction Booster (HIB) and Cross-modal Interaction Booster (CIB). In addition, VLN-PETL also incorporates vanilla adapters to efficiently tune the language encoder and LoRA to further improve the performance.

%% file: section/method.tex
\vspace{-5pt}
\section{Preliminaries}
\label{sec:methods}

\subsection{Problem Definition of VLN}
\vspace{-3pt}
Given a natural language instruction $\mathcal{I}$
and the current node on the connectivity graph of the environment, the VLN agent should predict an action $a_t$ at each time step $t$.
Specifically, at the time step $t$, the agent receives a panoramic view of the surrounding environment as the visual observation $\mathcal{O}_t$. $\mathcal{O}_t$ consists of $N$ single-view images and can be represented as a set of features: 
$\mathcal{O}_t=\{(v_n^o;a_n^o)\}$, where $v_n^o$ represents the feature of the $n$-th single-view image and $a_n^o$ represents the corresponding angle feature. The history $\mathcal{H}_t =\{(\mathcal{O}_i;a_i^h)\}$ that consists of every observation $\mathcal{O}_i$ and the performed action $a_i^h$ (\ie the turned angles) at each time step $i$ before $t$ also plays a vital role in the agent's action prediction.
Based on $\mathcal{I}$, $\mathcal{O}_t$ and $\mathcal{H}_t$, the VLN agent predicts the next action $a_t$ at each time step $t$ to navigate to the target goal until the special \texttt{[STOP]} action is selected, or the step reaches the maximum length. Some VLN tasks such as REVERIE \cite{reverie} additional require the agent to return a target object location, while some others such as the NDH \cite{ndh} use dialogue instructions.

\subsection{PETL Methods}

\noindent\textbf{Adapter} inserts trainable bottleneck layers after the multi-head attention layers or feed-forward layers in the transformer blocks. The bottleneck layer of an adapter consists of a linear down-projection with $\mW_{\text{down}} \in \sR^{D_{\text{hidden}} \times D_{\text{mid}}}$, a non-linear activation function $\sigma(\cdot)$ and a linear up-projection $\mW_{\text{up}} \in \sR^{D_{\text{mid}} \times D_{\text{hidden}}}$.  Given the input feature $\vf_\text{in}$, the adapter first projects $\vf_\text{in}$ into the $D_{\text{mid}}$ bottleneck dimension and then recovers it back into $D_{\text{hidden}}$ dimension as:
\begin{equation}
    \vf_\text{out}= \mW_{\text{up}}^\intercal \sigma(\mW_{\text{down}}^\intercal \vf_\text{in}). \label{con:eq1}
\end{equation}
Bias terms are omitted for brevity.
The parameters of layer normalization are usually tuned together with the adapter.
Besides, the adapter can be inserted into the transformer layers in a sequential manner or in a parallel manner, while the latter one is proved superior to the former one as in ~\cite{unified-petl}.

\noindent\textbf{LoRA} injects trainable low-rank decomposition matrices to represent the weight updates of the frozen parameters in the transformer's linear projection layers. Specifically, for a weight matrix $\mW\in\sR^{D_{\text{hidden}} \times D_{\text{hidden}}}$ in the pre-trained model, the weight update $\Delta\mW\in\sR^{D_{\text{hidden}} \times D_{\text{hidden}}}$ is approximated by two low-rank matrices $\mW_{\text{down}}\in\sR^{D_{\text{hidden}} \times D_{\text{mid}}}$ and $\mW_{\text{up}} \in \sR^{D_{\text{mid}} \times D_{\text{hidden}}}$ as follow:
\begin{equation}
   \mW + \Delta\mW = \mW + \mW_{\text{down}}\mW_{\text{up}} ,
\end{equation}
and the forward pass of LoRA can be formulated as:
\begin{equation}
    \vf_{\text{out}} = (\mW^\intercal + \gamma \mW_{\text{up}}^\intercal \mW_{\text{down}}^\intercal) \vf_{\text{in}}.
\end{equation}
where $\gamma$ is a fixed scalar hyperparameter for scaling.

\noindent\textbf{Prompt Tuning} prepends a sequence of randomly initialized continuous prompts $\vf_{\text{prompt}}$ into the input feature $\vf_\text{in}$. During training, only these prompts are optimized by updating a learnable projection matrix $\mW_\text{prompt}\in\sR^{1 \times D_{\text{hidden}}}$. %
The forward pass can be formulated as:
\begin{align}
\vf_{\text{prompt}} &= \mW_\text{prompt}^\intercal\mX, \\
\vf_{\text{out}} &= \text{TRM}([\vf_{\text{prompt}},\vf_{\text{in}}]).
\end{align}
where $\mX$ represents discrete prompt tokens, $[\cdot]$ represents concatenation, $\text{TRM}(\cdot)$ represents the transformer block.

\noindent\textbf{BitFit} does not introduce new parameters or inputs for tuning. It only tunes the bias terms in the pretrained models, which shows promising performance in some NLP tasks. %

\section{Method}
\vspace{-3pt}
\subsection{Baseline Model}
\label{sec:transformer_model}
\begin{figure}[!t]
	\begin{center}
		\includegraphics[width=\linewidth]{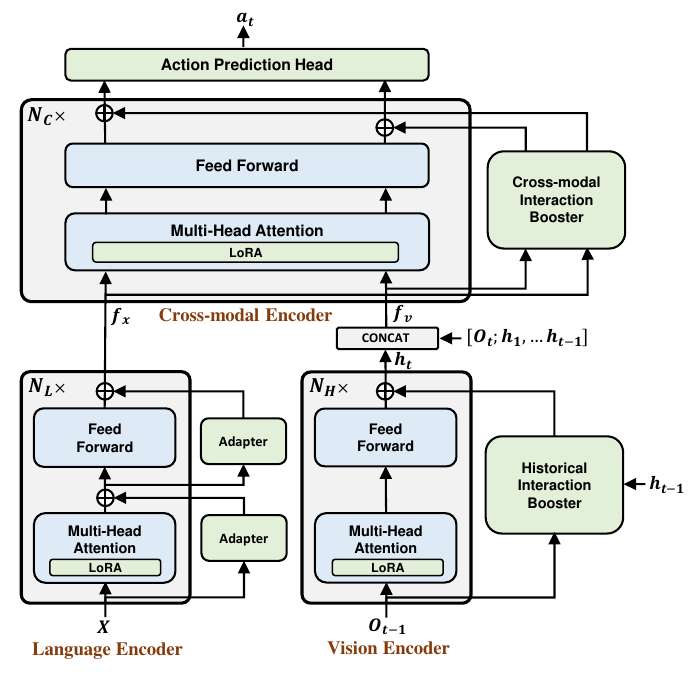}
	\end{center}
	\vspace{-15pt}
	\caption{Illustration of the framework of our proposed VLN-PETL. The pre-trained model of HAMT mainly consists of a language encoder, a vision encoder and a cross-modal encoder. The blue color denotes the frozen parameters in the pre-trained model and the green color denotes the trainable parameters of injected PETL modules.}
	\label{fig:vln}
	\vspace{-10pt}
\end{figure}

\vspace{-5pt}
We use a large pre-trained model of HAMT~\cite{hamt} as our baseline, which is pre-trained on plentiful text-image pairs of instructions and the corresponding trajectories.
As illustrated in Figure~\ref{fig:vln}, HAMT adopts a two-stream architecture, which consists of a language encoder and a vision encoder to extract single-modal features, and a cross-modal encoder to fuse multi-modal features for action prediction.  %

\noindent\textbf{Language Encoder} Following the practice of BERT, the language encoder first embeds the instruction $\mathcal{I}$ into language embedding $E_{x}$ by summing the word embedding, position embedding, and type embedding of each word $x_i$ in the instruction. Then, $E_{x}$ is passed through $N_L$ transformer blocks which consist of a multi-head self-attention layer and feed-forward layer to generate the language feature $\vf_x$.


\noindent\textbf{Vision Encoder} 
The vision encoder mainly consists of observation encoding and history encoding. At time step $t$, the panoramic image feature $v_{t}^o$ 
and the corresponding angle feature $a_{t}^o$ are projected into $E_v^o$ and $E_a^o$ followed by the layer normalization. Then, $E_v^o$ and $E_a^o$ are summed up as the current observation feature $\vo_t$. Meanwhile, the panoramic image feature $v^h_{t-1}$ and the corresponding turned angle feature $a_{t-1}^h$ of the previous time step are taken as the input for history encoding. Similarly, $v^h_{t-1}$ and $a_{t-1}^h$ are first projected into $E_v^h$ and $E_a^h$. Then, $E_v^h$ and $E_a^h$ are summed up and passed through $N_H$ transformer blocks to generate $h_t$, which is appended into the tail of $\langle h_1,...,h_{t-1}\rangle$ as the current time step's history feature $\vh_t$. Indeed, the history encoding only re-encodes the previous time step's observation statically without any other interactions. Thus, this manner of history feature's generation and update may neglect the temporal knowledge embodied in the navigation trajectory. 

\noindent\textbf{Cross-modal Encoder}
At the time step $t$, the observation feature $\vo_t$ and the history feature $\vh_t$
are first concatenated as the visual feature $\vf_v$. Then, $\vf_x$ and $\vf_v$ are passed through $N_C$ cross-modal transformer blocks that consist of two-stream multi-head cross-attention layers, multi-head self-attention layers, and feed-forward layers to generate cross-modal features for action prediction.

\vspace{-5pt}
\subsection{VLN-PETL}
\vspace{-3pt}
Not all the aforementioned PETL methods perform well in the complicated VLN tasks, such as BitFit and Prompt Tuning, which bring performance degeneration to VLN tasks (see Sec.~\ref{sec:results} for evaluation and explanation).  %
Thus, as shown in Figure~\ref{fig:vln}, we only incorporate the adapter and LoRA as the PETL components of our integrated VLN-PETL. 
Furthermore, two block-level PETL modules are specially designed for VLN tasks considering the unique characteristics of VLN tasks and parameter efficiency, namely Historical Interaction Booster (HIB) and Cross-modal Interaction Booster (CIB). Based on the bottleneck structure of the adapter, these two modules respectively strengthen the historical interaction and cross-modal interaction by incorporating the multi-head cross-attention mechanism and gating mechanism. 

\begin{figure*}[!ht]
  \centering
  \begin{subfigure}{0.33\linewidth}
    \centering
    \parbox[][5.5cm][c]{\linewidth}{
    \includegraphics[scale=0.48]{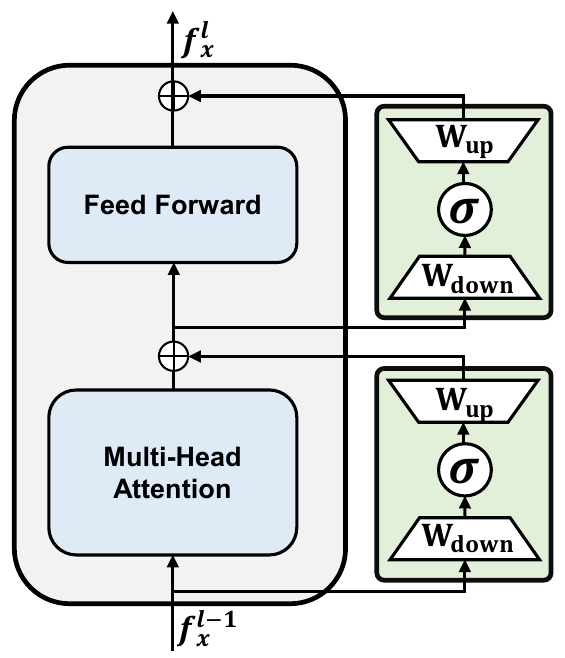}
    }
    \vspace{-2pt}
    \caption{Language Encoder Adapter (LEA).}
    \label{fig:lang}
  \end{subfigure}
  \hfill
  \begin{subfigure}{0.33\linewidth}
    \centering
    \parbox[][5.5cm][c]{\linewidth}{
    \includegraphics[scale=0.47]{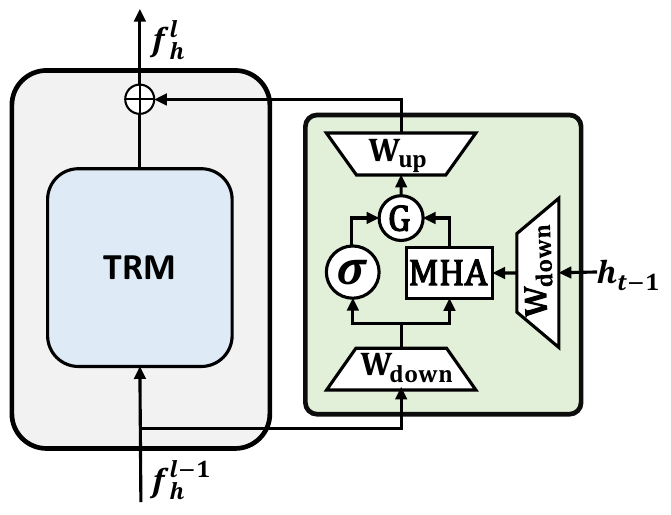}
    }
    \vspace{-2pt}
    \caption{Historical Interaction Booster(HIB).}
    \label{fig:hib}
  \end{subfigure}
  \hfill
  \begin{subfigure}{0.33\linewidth}
    \centering
    \parbox[][5.5cm][c]{\linewidth}{
    \includegraphics[scale=0.47]{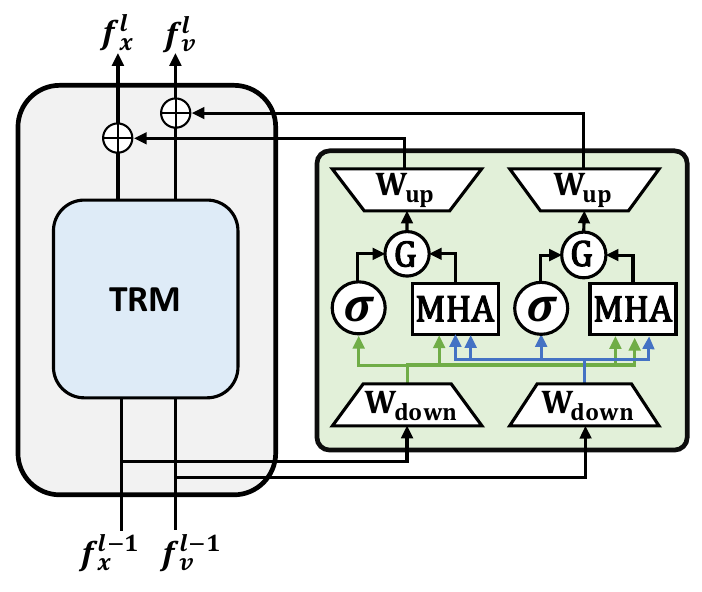}
    }
    \vspace{-2pt}
    \caption{Cross-modal Interaction Booster(CIB).}
    \label{fig:cib}
  \end{subfigure}
  \vspace{-15pt}
  \caption{Detailed components of VLN-PETL. TRM represents the transformer block, MHA represents the multi-head attention layer, $\sigma$ represents the activation layer and $G$ represents the learnable gate.}
  \label{fig:petl}
  \vspace{-15pt}
\end{figure*}

\noindent\textbf{Language Encoder Adapter}
As shown in Figure~\ref{fig:lang}, the Language Encoder Adapter (LEA) is inserted into the multi-head self-attention layers and feed-forward layers in parallel. 
Concretely, for the $l$-th transformer block in the language encoder, the input feature $\vf^{l-1}_x$ is first passed through the adapter 
and summed with the output feature of the multi-head self-attention layer as $\vf_{\text{att}}$:
\begin{align}
    \vf_{\text{att}} &= \text{LN}\big(\text{MSA}(\vf^{l-1}_x) + \text{ADAPTER}(\vf^{l-1}_x)\big),
\end{align}
where $\text{MSA}(\cdot)$ represents the multi-head self-attention layer, $\text{ADAPTER}(\cdot)$ represents the adapter block shown in Eq.\ref{con:eq1} and $\text{LN}(\cdot)$ represents the layer normalization. 
Similarly, another adapter is inserted into the feed-forward layer which takes $\vf_{\text{att}}$ as input:
\begin{align}
    \vf_{\text{ffn}} &= \text{LN}\big(\text{FFN}(\vf_{\text{att}}) + \text{ADAPTER}(\vf_{\text{att}})\big),
\end{align}
where $\text{FFN}(\cdot)$ represents the feed-forward layer. $\vf_{\text{ffn}}$ is used as the final output feature $\vf^{l}_x$ of the $l$-th transformer block:
\begin{align}
\vf^{l}_x &= \vf_{\text{ffn}}.
\end{align}

\noindent\textbf{Historical Interaction Booster}
As shown in Figure~\ref{fig:hib}, the Historical Interaction Booster (HIB) adopts the multi-head cross-attention mechanism to enhance the historical interaction between the observation feature and the history feature at each timestep $t$ in a recurrent pattern.
Specifically, for the $l$-th transformer block in the vision encoder, the input observation feature $\vf_h^{l-1}$ and the previous history feature $\vh_{t-1}$ are first downsampled into $\vf_\text{down}$ and $\vh_\text{down}$ with $D_\text{mid}$ dimension by projection matrices $\mW_{\text{down\_f}}$ and $\mW_{\text{down\_h}}$:
\begin{align} 
\vf_\text{down} &= \mW_{\text{down\_f}}^\intercal \vf_{h}^{l-1}, \label{con:eq5} \\
\vh_\text{down} &= \mW_{\text{down\_h}}^\intercal \vh_{t-1}, \label{con:eq6}
\end{align}
Then, the history knowledge is encoded into the observation feature by the multi-head cross-attention between $\vf_\text{down}$ and $\vh_\text{down}$ followed by a learnable gate $\alpha$: 
\begin{align} 
\vf'_\text{down} &= \text{ReLU}(\vf_\text{down}), \label{con:eq7} \\
\vf_\text{cross} &= \text{MHA}(\vf_\text{down}, \vh_\text{down}), \\
\alpha &= \text{Sigmoid}(\frac{\theta}{T}), \label{con:eq8} \\
\vf_\text{v\_h} &= \alpha\ast\vf'_\text{down} + (1-\alpha)\ast\vf_\text{cross},
\end{align}
where $\text{MHA}(\cdot)$ represents the multi-head cross-attention layer of which the query is $\vf_\text{down}$ while the key and value are $\vh_\text{down}$, $\theta$ is a learnable scalar initialized by zero and $T$ is fixed as 0.1 representing the temperature hyperparameter. 
Next, the attended visual-and-historical feature $\vf_\text{v\_h}$ is upsampled into $D_\text{hidden}$ dimension and summed with the original output feature ${\hat{\vf}}_{t}^{l}$ of the $l$-th transformer block:
\begin{align} 
\hat{{\vf}}_{h}^{l} &= \text{TRM}(\vf_{h}^{l-1}), \\
\vf_{h}^{l} &= \text{LN}(\hat{{\vf}}_{h}^{l} + \mW_{\text{up}}^\intercal\vf_\text{v\_h}),
\end{align}
where $\text{TRM}(\cdot )$ represents the transformer block in the vision encoder. 
The final output history feature $h_{t}$ at the current timestep $t$ is obtained as follow:
\begin{align} 
h_{t} = \vf_{h}^{L},
\end{align}
where $L$ represents the number of transformer blocks.

\noindent\textbf{Cross-modal Interaction Booster}
As shown in Figure~\ref{fig:cib}, to enhance the interaction between language and visual modalities, Cross-modal Interaction Booster (CIB) adopts a two-stream multi-head cross-attention mechanism. 
To be specific, for the $l$-th transformer block in the cross-modal encoder, the input language feature $\vf_x^{l-1}$ and the visual feature $\vf_v^{l-1}$ are first downsampled into  $\vf_\text{down\_x}$ and $\vf_\text{down\_v}$ with $D_\text{mid}$ dimension by projection matrices $\mW_\text{down\_x}$ and $\mW_\text{down\_v}$ as Eq.\ref{con:eq5} and Eq.\ref{con:eq6}. Then, a two-stream multi-head cross-attention is implemented by exchanging the query for the key and value as follows:
\begin{align}
\vf_\text{cross\_x} &= \text{MHA}(\vf_\text{down\_x}, \vf_\text{down\_v}), \\
\vf_\text{cross\_v} &= \text{MHA}(\vf_\text{down\_v}, \vf_\text{down\_x}), 
\end{align}
Two learnable gates $\alpha_x$ and $\alpha_v$ are used to obtain cross-attended language feature $\vf_\text{x\_v}$ and visual feature $\vf_\text{v\_x}$: 
\begin{align} 
\vf_\text{x\_v} &= \alpha_x\ast\vf'_\text{down\_x} + (1-\alpha_x)\ast\vf_\text{cross\_x}, \\
\vf_\text{v\_x} &= \alpha_v\ast\vf'_\text{down\_v} + (1-\alpha_v)\ast\vf_\text{cross\_v},
\end{align}
where $\vf'_\text{down\_x}$ and $\vf'_\text{down\_v}$ are obtained by passing $\vf_\text{down\_x}$ and $\vf_\text{down\_v}$ through ReLU layer as Eq.\ref{con:eq7}. 
At last, $\vf_\text{x\_v}$ and $\vf_\text{v\_x}$ are respectively upsampled into $D_\text{hidden}$ dimension and summed with the original output language feature ${\hat{\vf}}_x^{l}$ and visual feature ${\hat{\vf}}_v^{l}$ as the final output feature $\vf_{x}^{l}$ and $\vf_{v}^{l}$ of the $l$-th transformer block:
\begin{align} 
\hat{{\vf}}_{x}^{l},\hat{{\vf}}_{v}^{l} &= \text{TRM}(\vf_{x}^{l-1}, \vf_{v}^{l-1}), \\
\vf_{x}^{l} &= \text{LN}(\hat{{\vf}}_{x}^{l} + \mW_{\text{up\_x}}^\intercal\vf_\text{x\_v}), \\
\vf_{v}^{l} &= \text{LN}(\hat{{\vf}}_{v}^{l} + \mW_{\text{up\_v}}^\intercal\vf_\text{v\_x}).
\end{align}
where $\text{TRM}(\cdot )$ represents the transformer block in the cross-modal encoder.

\begin{figure}[!t]
	\begin{center}
		\includegraphics[width=0.6\linewidth]{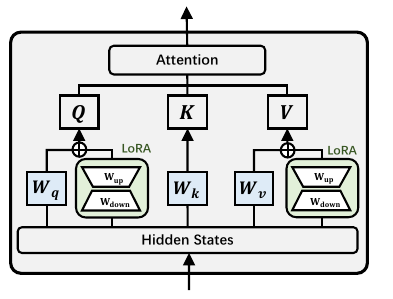}
	\end{center}
	\vspace{-18pt}
	\caption{Illustration of injecting LoRA in the Multi-head Attention layer.}
	\label{fig:lora}
	\vspace{-16pt}
\end{figure}

\noindent\textbf{Incorporating LoRA in VLN-PETL}
The multi-head attention layers account for a large portion of parameters in the transformer. 
Thus, 
as shown in Figure~\ref{fig:lora}, VLN-PETL incorporates LoRA as an independent PETL component in addition to LEA, HIB and CIB to further improve the performance on downstream VLN tasks. Specifically, we globally inject LoRA layers into query matrices $\mW_Q$ and value matrices $\mW_V$ of all multi-head attention layers  
in the pre-trained model. Given the input feature $\vf_{\text{in\_q}}$ and $\vf_{\text{in\_v}}$ for linear projection in the multi-head attention layer, the output feature $\mQ$ and $\mV$ can be computed as follow:
\begin{align}
    \mQ &= (\mW_Q^\intercal + \gamma \mW_{\text{up\_q}}^\intercal \mW_{\text{down\_q}}^\intercal) \vf_{\text{in\_q}}, \\
    \mV &= (\mW_V^\intercal + \gamma  \mW_{\text{up\_v}}^\intercal \mW_{\text{down\_v}}^\intercal) \vf_{\text{in\_v}}.
\end{align}

%% file: section/experiments.tex
\section{Experimental Setup}
\label{sec:experiment}

\subsection{Downstream tasks}
To comprehensively evaluate PETL methods for VLN, %
we conduct experiments on four downstream tasks: Room-to-Room (R2R)~\cite{r2r}, REVERIE~\cite{reverie}, NDH~\cite{ndh} and Room-across-Room (RxR)~\cite{rxr}. These downstream tasks evaluate the agent from different views: (1) R2R and RxR require the agents to follow detailed instructions to navigate from one room to another; (2) REVERIE gives a concise, high-level instruction referring to a remote object, which focuses on grounding remote target objects; (3) NDH requires an agent to reach target regions based on the dialog history, which contains multiple question-and-answer interactions between the agent and an oracle. 

\input{section/table_r2r}
\input{section/table_reverie}

\begin{table}[!t]
\centering
\small
\tabcolsep=0.07cm
\resizebox{0.85\linewidth}{!}{
\begin{tabular}{lcccc} \toprule
\multicolumn{1}{l}{\multirow{2}{*}{Methods}} &Updated& \multirow{2}{*}{Val Seen} &  \multirow{2}{*}{Val Unseen} &\multirow{2}{*}{Test Unseen} \\ 
~ &Params(\%) & ~ & ~ & ~\\ \midrule
Fine-Tuning & 100 & 7.69 & 5.16 &5.05\\
\midrule
BitFit~\cite{BenZaken2022BitFitSP}&0.46&\underline{6.68}&3.77&4.03\\
Prompt Tuning~\cite{lester-etal-2021-power}&0.37&4.71&3.26&2.86\\
LoRA~\cite{hu2021lora}&3.02&6.83&5.16&\underline{5.91}\\
Adapter~\cite{houlsby2019adapter}&3.08& 5.29 &\underline{5.30}&4.72\\
\midrule
VLN-PETL(ours)&2.82&\textbf{7.76}&\textbf{5.69}&\textbf{6.13}\\
\bottomrule
\end{tabular}
}
\vspace{-1mm}
\caption{Performance on NDH measured by Goal Progress.}
\vspace{-20pt}
\label{tab:CVDN_sota_cmpr}
\end{table}

\begin{table*}[!t]
\centering
\resizebox{0.8\linewidth}{!}{
\begin{tabular}{l ccccc|cccc}
\toprule
\multicolumn{1}{l}{\multirow{2}{*}{Methods}} &Updated& \multicolumn{4}{c}{RxR Validation Seen} & \multicolumn{4}{c}{RxR Validation Unseen} \\ 
~ &Params(\%) & SR$\uparrow$ & SPL $\uparrow$ & nDTW $\uparrow$ & sDTW $\uparrow$ & SR$\uparrow$ & SPL $\uparrow$ & nDTW $\uparrow$ & sDTW$\uparrow$ \\ 
\midrule
Fine-Tuning & 100 &64.93&61.28&69.26&55.92 & 57.88 &54.18&64.52&49.44\\
\midrule
BitFit~\cite{BenZaken2022BitFitSP}&0.28&35.69&33.58&49.37&29.81&36.63&34.34&50.47&30.49\\
Prompt Tuning~\cite{lester-etal-2021-power}&0.23&27.92&25.98&42.93&22.97&29.95&27.72&44.94&24.67\\
LoRA~\cite{hu2021lora}&1.86&55.66&52.41&63.20&47.31&54.53&51.14&63.23&46.94\\
Adapter~\cite{houlsby2019adapter}&1.90&\underline{58.52}&\underline{54.96}&\underline{65.14}&\underline{50.23}&\underline{55.19}&\underline{51.44}&\underline{63.56}&\underline{47.32}\\
\midrule
VLN-PETL(ours)& 1.67&\textbf{60.48}&\textbf{56.77}&\textbf{65.74}&\textbf{51.67}&\textbf{57.95}&\textbf{54.16}&\textbf{64.94}&\textbf{49.70}\\
\bottomrule
\end{tabular}}
\vspace{-2mm}
\caption{Performance on RxR using English instructions. nDTW is the main metric for the RxR task.
}
\vspace{-10pt}
\label{tab:main_result_rxr}
\end{table*}

\begin{table}[t]
\vspace{-0pt}
\centering
\renewcommand{\arraystretch}{0.3}
\resizebox{\linewidth}{!}{
\begin{tabular}{l |ccc |c |cc |cc}
\toprule
&\multicolumn{4}{c}{Components} & \multicolumn{2}{c}{REVERIE Val Unseen}&\multicolumn{2}{c}{RxR Val Unseen}\\
&LEA & HIB& CIB& LoRA & \multicolumn{1}{c}{SPL$\uparrow$} & \multicolumn{1}{c|}{RGSPL$\uparrow$}& SR$\uparrow$ & SPL$\uparrow$\\
\midrule
1&$\times$ &$\times$&$\times$&$\times$&15.19&4.53&28.66&26.72\\
2&\checkmark&$\times$  &$\times$& $\times$ &18.21&9.73&43.45&40.68 \\
3&$\times$& \checkmark &$\times$ &$\times$&18.49&9.32&41.04&38.46\\
4&$\times$& $\times$& \checkmark&$\times$&21.62&12.86&52.01&48.82 \\
5&\checkmark& \checkmark& $\times$& $\times$&21.86  & 11.75 &45.08&42.11\\
6&\checkmark& $\times$& \checkmark& $\times$&25.00& 14.34 &52.26&49.26\\
7&$\times$& \checkmark& \checkmark& $\times$&25.55&14.89&54.26&50.71  \\
8&\checkmark& \checkmark& \checkmark& $\times$&26.51&15.29&56.04&52.79\\
9&\checkmark& \checkmark& \checkmark&\checkmark&\textbf{27.67}&\textbf{15.96}&\textbf{57.95}&\textbf{54.16}\\
\bottomrule
\end{tabular}}
\vspace{-2mm}
\caption{Ablation of different components in VLN-PETL on REVERIE Unseen set and RxR Unseen set.}
\label{tab:abla_comp}
\vspace{-5pt}
\end{table}

\begin{table}[!t]
\centering
\renewcommand{\arraystretch}{0.5}
\resizebox{0.8\linewidth}{!}{
\begin{tabular}{c cc| cc}
\toprule
\multicolumn{1}{c}{\multirow{2}{*}{Methods}} & \multicolumn{2}{c}{Validation Seen}&\multicolumn{2}{c}{Validation Unseen}\\
~& \multicolumn{1}{c}{SPL$\uparrow$} & \multicolumn{1}{c}{RGSPL$\uparrow$}& SPL$\uparrow$ & RGSPL$\uparrow$\\
\midrule
HEA&25.74&12.51&18.98&8.39\\
HIB&28.55&16.95&18.49&9.32\\
\midrule
CEA&31.44&17.55&20.77&10.14\\
CIB&38.99&25.46&21.62&12.86\\
\bottomrule
\end{tabular}}
\vspace{-3mm}
\caption{Performance comparison of HIB and CIB with their counterparts of Adapter on REVERIE val set. 
}
\label{tab:abla_repre}
\end{table}

\begin{table}[tbp]
\vspace{-3pt}
\centering
\resizebox{0.85\linewidth}{!}{
\begin{tabular}{c cc| cc}
\toprule
\multicolumn{1}{c}{\multirow{2}{*}{Methods}} & \multicolumn{2}{c}{Validation Seen}&\multicolumn{2}{c}{Validation Unseen}\\
~& \multicolumn{1}{c}{SPL$\uparrow$} & \multicolumn{1}{c}{RGSPL$\uparrow$}& SPL$\uparrow$ & RGSPL$\uparrow$\\
\midrule
VLN-PETL&42.60&27.61&27.67&15.96\\
w/o LoRA&41.34&27.12&26.51&15.29\\
\bottomrule
\vspace{-8mm}
\end{tabular}}
\caption{Ablation of LoRA's effect on REVERIE val set.}
\label{tab:lora}
\end{table}

\begin{table}[t]
\vspace{0pt}
    \centering
	\begin{minipage}{0.495\linewidth}
		\centering
          \vspace{-1mm}
          \renewcommand{\arraystretch}{0.6}
          \setlength\tabcolsep{2pt}
        \label{table:NormalSafe}
        \resizebox{0.85\linewidth}{!}{
        \begin{tabular}{lcccc}
        \toprule
		$T$ & 0.01 & 0.1 & 1 & 10\\
		\midrule
		  SPL$\uparrow$ &25.60 &27.67 &\textbf{27.81}&25.00 \\
    	RGSPL$\uparrow$ & 14.41& \textbf{15.96} &15.36&14.06 \\
            \bottomrule
        \end{tabular}}
	\end{minipage}
	\hfill
	\begin{minipage}{0.45\linewidth}
		\centering
          \vspace{-1mm}
          \renewcommand{\arraystretch}{0.9}
          \setlength\tabcolsep{14pt}
        \label{table:NormalSafe}
        \resizebox{1.0\linewidth}{!}{
        \begin{tabular}{lcc}
        \toprule
		$\alpha$ & 0.5 & learnable \\
		\midrule
		  SPL$\uparrow$ &27.51 &\textbf{27.67}\\
    	RGSPL$\uparrow$ & 14.50&\textbf{15.96} \\
            \bottomrule
        \end{tabular}}
	    \end{minipage}
     \vspace{-7pt}
     \caption{Ablation of $T$ and $\alpha$ in the gates of HIB and CIB on REVERIE Val Unseen set.}
     \label{tab:T and gate}
     \vspace{-15pt}
\end{table}

\subsection{Evaluation metrics}
We follow previous work and adopt the most commonly used metrics for evaluating VLN agents as follows: \textbf{\texttt{TL}} (Trajectory Length) measures the average length of all the predicted navigation trajectories in meters. 
\textbf{\texttt{NE}} (Navigation Error) is the mean the average distance in meters between the agent's final location and the target location. 
\textbf{\texttt{SR}} (Success Rate) measures the ratio of successful tasks, of which the agent's stop location is less than 3 meters away from the target location.
\textbf{\texttt{SPL}} (Success weighted by Path Length~\cite{spl}) trades-off SR (Success Rate) against TL (Trajectory Length), which measures both the accuracy and efficiency of navigation. 
\textbf{\texttt{OSR}} (Oracle Success Rate) measures the ratio of tasks of which one of its trajectory viewpoints can observe the target object within 3 meters. 
\textbf{\texttt{RGS}} (Remote Grounding Success rate) measures the ratio of tasks that successfully locate the target object.
\textbf{\texttt{RGSPL}} (RGS weighted by Path Length) is RGS weighted by Path Length.
\textbf{\texttt{GP}} (Goal Progress) measures the average progress of the agent towards the target.
\textbf{\texttt{nDTW}} (Normalized Dynamic
Time Warping) penalizes deviations from the reference path.
\textbf{\texttt{sDTW}} (Success weighted by normalized Dynamic TimeWarping) constrains nDTW to only successful episodes and effectively captures both success and fidelity.

\subsection{Implementation details}
We choose existing PETL methods of BitFit, Prompt Tuning, Adapter and LoRA for comparison with our integrated VLN-PETL. We use the same learning rate of $1e-4$, and AdamW~\cite{adamw} optimizer for all PETL methods. The batch size is set as 4 for REVERIE and 8 for the other three VLN tasks. For Prompt Tuning, we respectively add 20 prompt tokens in front of the inputs of the language encoder and vision encoder. For the setting of both Adapter and LoRA, the bottleneck dimension $D_\text{dim}$ is set as 64, which brings comparative parameters for a fair comparison with VLN-PETL. While for VLN-PETL, the bottleneck dimensions $D_\text{dim}$ of the Language Encoder Adapter (LEA), History Interaction Booster (HIB), and Cross-modal Interaction Booster (CIB) is set as 64 while the bottleneck dimension of the incorporated LoRA is set as 8 for REVERIE while 16 for other VLN tasks. The attention heads number of both HIB and CIB is set as 4. For all PETL methods, the prediction heads of the pre-trained model are also trained. 
For a fair comparison, we follow HAMT~\cite{hamt} to use Reinforcement Learning (RL) and Imitation Learning (IL).
\section{Experimental Results}
\label{sec:results}

\subsection{Comparison of PETL methods for VLN}
As shown in Table~\ref{tab:main_result_r2r}-\ref{tab:main_result_rxr}, we compare our proposed VLN-PETL with finetuning, Bitfit~\cite{BenZaken2022BitFitSP}, Prompt Tuning~\cite{lester-etal-2021-power}, LoRA~\cite{hu2021lora}, and Adapter~\cite{houlsby2019adapter} in performance and trainable parameter amounts on different VLN tasks.

We can see that Prompt Tuning with the least updated parameters works poorly for all VLN downstream tasks. One possible reason may be the limited trainable parameters. More importantly, the training instability of Prompt Tuning as previous work declared should also be responsible for the poor performance. In fact, training VLN agents itself is not always stable and easy, where reinforcement learning plays a vital role. BitFit surpasses Prompt Tuning with a non-trivial margin on all tasks with comparable amounts of trainable parameters. However, the performance of BitFit still falls far from finetuning, LoRA, and Adapter on all VLN tasks especially on RxR in high demand for language understanding with much longer instructions. Only tuning bias terms may have difficulties in handling these complex VLN tasks. Thus, we believe that Prompt Tuning and BitFit are not applicable for efficiently tuning large pre-trained models for challenging VLN tasks. 
While LoRA and Adapter not only have comparative amounts of trainable parameters but also have comparable performances on all VLN tasks. These two methods further shrink the performance gap with finetuning, which are potential to effectively tune VLN pretrained models.

As for VLN-PETL, though it has fewer parameters than LoRA and Adapter, VLN-PETL still surpasses LoRA and Adapter on most evaluation metrics in all four downstream VLN tasks. Furthermore, only VLN-PETL maintains competitive performances compared to fine-tuning and even outperforms fine-tuning on several evaluation metrics. As shown in Table~\ref{tab:CVDN_sota_cmpr}, it is worth mentioning that VLN-PETL outperforms full fine-tuning on all dataset splits in the NDH task, and achieves the top position on the public leaderboard\footnote{\url{https://eval.ai/web/challenges/challenge-page/463/leaderboard/1292} (01/03/2023)}. These promising results demonstrate the effectiveness of our proposed VLN-PETL for efficiently tune large pre-trained models for VLN tasks.%

\subsection{Ablation Study}
\paragraph{Contribution of VLN-PETL components}
As shown in Table~\ref{tab:abla_comp}, to evaluate the contribution of LEA, HIB and CIB, we choose REVERIE and RxR which are more challenging VLN tasks to conduct ablation studies. REVERIE not only measures the agent's ability in navigation but also in locating the target object, while RxR has much longer instructions requiring comprehensive language understanding. We also report the results of only tuning the prediction head for comparison. We find that LEA has a competitive performance compared to HIB, and only tuning either LEA or HIB outperforms the head tuning with a nontrivial margin. While CIB contributes much more inefficiently tuning the VLN model, which improves the performance with a larger margin on both REVERIE and RxR. This result indicates that language understanding and vision understanding with history knowledge contribute comparably to the action prediction for VLN agents, while the cross-modal interaction plays more importantly in this process.
By combining all these three components, the VLN agent achieves a promising performance and outperforms other PETL methods, especially on the RGSPL metric, which measures the agent's ability to locate the target object. 

\vspace{-16pt}
\paragraph{Superiority of HIB and CIB}
As shown in Table~\ref{tab:abla_repre}, to validate the effectiveness of HIB and CIB, we compare the performance of HIB and CIB with their counterparts of Adapter, by respectively replacing HIB and CIB by History Encoder Adapter (HEA) and Cross-modal Encoder Adapter (CEA) which are similar to Language Encoder Adapter. %
Due to the enhancement of historical knowledge learning, HIB surpasses HEA on seen set by a large margin. On the unseen set, HIB falls behind HEA with a trivial margin on the SPL metric while outperforming HEA on the RGSPL metric with a large margin. This is probably because the input for history encoding is a panoramic view image rather than a single-view image of the front view, where HIB tends to learn more knowledge about the fine-grained object rather than the trajectory. %
As for CIB and CEA, CIB surpasses CEA on all metrics and all sets with a large margin, which shows the superiority of CIB.

\paragraph{The Effect of LoRA}
As shown in Table~\ref{tab:lora}, we find that when removing LoRA, the performance of VLN-PETL has a slight drop on all main metrics on both REVERIE seen and unseen splits. Besides, the decrease on RGSPL metric is less than that on SPL metric, which indicates LoRA's effect on the VLN agent's ability to locate objects is smaller than that of navigation during efficient tuning.

\vspace{-8pt}
\paragraph{Hyper-parameters in Gates.}
As shown in Table~\ref{tab:T and gate}, we conduct ablation studies on $T$ and $\alpha$ in the gates of HIB and CIB. The performances are similarly high when $T$ is set as 0.1 or 1. We set $T$ as 0.1 due to its higher score on RGSPL. We also compare the results of fixing $\alpha$ as 0.5 and using the learnable gate. We can see that the learnable gate $\alpha$ surpasses the fixed $\alpha$ with a large margin on RGSPL. 

%% file: section/table_r2r.tex
\begin{table*}[!t]
\centering
\vspace{-2mm}
\resizebox{0.8\linewidth}{!}{
\begin{tabular}{lcccccccccccccc}
\toprule
\multicolumn{1}{l}{\multirow{2}{*}{Methods}}& Updated &\multicolumn{4}{c}{Validation Seen} & \multicolumn{4}{c}{Validation Unseen} & \multicolumn{4}{c}{Test Unseen} \\ 
~ & Params(\%) & TL & NE $\downarrow$ & SR $\uparrow$ & SPL $\uparrow$ & TL & NE $\downarrow$ & SR $\uparrow$ & SPL$\uparrow$  & TL & NE $\downarrow$ & SR $\uparrow$ & SPL $\uparrow$\\ 
\midrule
Fine-Tuning & 100 & 11.48 & 2.94 & 72.67 & 69.17 & 11.62 & 3.64 & 64.24& 59.25 & 12.20 & 4.09 & 63.20 & 58.55\\
\midrule
BitFit~\cite{BenZaken2022BitFitSP}&0.46 &11.61 &3.78 &63.47 &60.35 & 12.22&4.18&59.17&54.67&12.96&4.63&57.15&53.03\\
Prompt Tuning~\cite{lester-etal-2021-power}&0.37&10.67&4.24&61.02&58.59&11.14&4.63&56.49&52.32&11.60&4.88&54.47&50.91\\
LoRA~\cite{hu2021lora}&3.02&11.73 & \underline{3.14} & \underline{70.13} & \underline{66.00} & 12.25 & \underline{3.84} & \underline{63.60} & \underline{57.59}&12.99&\underline{4.15}&\underline{61.44}&\underline{55.96}\\
Adapter~\cite{houlsby2019adapter}&3.08&11.70&3.34&67.38&64.42&12.66&4.00&63.01&57.42&13.19&4.27&60.69&55.88\\
\midrule
VLN-PETL(ours)&2.82&11.39&\textbf{2.93}&\textbf{72.28}&\textbf{68.50}&11.52&\textbf{3.53}&\textbf{65.47}&\textbf{60.01}&12.30&\textbf{4.10}&\textbf{63.22}&\textbf{58.25}\\
\bottomrule
\end{tabular}}
\vspace{-1mm}
\caption{
Performance of PETL methods on R2R. 
For each method, we report the percentage of trainable parameters compared to full fine-tuning. \textbf{Bold} and \underline{underline} denote the best and runner-up results. SPL is the main metric.
}
\vspace{-2pt}
\label{tab:main_result_r2r}
\end{table*}

%% file: section/table_reverie.tex
\begin{table*}[!t]
\centering
\vspace{-0mm}
\resizebox{1\linewidth}{!}{
\begin{tabular}{l c cccccc|cccccc|cccccc}
\toprule
\multicolumn{1}{l}{\multirow{3}{*}{Methods}} &Updated& \multicolumn{6}{c}{REVERIE Validation Seen} &\multicolumn{6}{c}{REVERIE Validation Unseen} & \multicolumn{6}{c}{REVERIE Test Unseen} \Tstrut\\
\cline{3-20}
~&~& \multicolumn{4}{c}{Navigation}  &  \multicolumn{1}{c}{\multirow{2}{*}{RGS$\uparrow$}}&
\multicolumn{1}{c|}{\multirow{2}{*}{RGSPL$\uparrow$}} & \multicolumn{4}{c}{Navigation}  & \multicolumn{1}{c}{\multirow{2}{*}{RGS$\uparrow$}}&
\multicolumn{1}{c|}{\multirow{2}{*}{RGSPL$\uparrow$}} & \multicolumn{4}{c}{Navigation}   & \multicolumn{1}{c}{\multirow{2}{*}{RGS$\uparrow$}}&
\multicolumn{1}{c}{\multirow{2}{*}{RGSPL$\uparrow$}}  \\
~&  Params(\%) &\multicolumn{1}{c}{SR$\uparrow$} & \multicolumn{1}{c}{OSR$\uparrow$} & \multicolumn{1}{c}{SPL$\uparrow$}  & \multicolumn{1}{c}{TL} &  &  & \multicolumn{1}{c}{SR$\uparrow$} & \multicolumn{1}{c}{OSR$\uparrow$} & \multicolumn{1}{c}{SPL$\uparrow$} &\multicolumn{1}{c}{TL} & & & \multicolumn{1}{c}{SR$\uparrow$} & \multicolumn{1}{c}{OSR$\uparrow$} & \multicolumn{1}{c}{SPL$\uparrow$} & \multicolumn{1}{c}{TL} &  & \Tstrut\\
\midrule
Fine-Tuning&100&46.73&52.35&42.75&13.37&30.64&27.91&32.63&37.82&28.92&15.66&18.66&16.06&33.09&37.82&27.02&12.83&15.04&13.32\\
\midrule
BitFit~\cite{BenZaken2022BitFitSP}&0.80&33.52&37.81&31.84&11.45&19.96&18.93&24.65&27.46&21.34&12.36&10.85&9.43&21.50&24.95&18.85&12.53&9.87&8.62\\
Prompt Tuning~\cite{lester-etal-2021-power}&0.71&25.86&33.17&23.21&12.15&8.29&7.44&19.94&25.08&17.75&12.43&5.82&5.07&19.95&24.24&17.94&11.61&5.51&4.88\\
LoRA~\cite{hu2021lora}&3.33&\underline{42.30}&46.24&\underline{38.63}&12.89&\underline{29.87}&\underline{27.45}&29.42&34.28&26.17&15.96&15.25&13.45&\textbf{32.12}&37.00&\textbf{26.86}&14.93&\underline{14.94}&\underline{12.76}\\
Adapter~\cite{houlsby2019adapter}&3.39&40.76&45.05&37.43&13.75&27.13&24.62&\underline{29.48}&32.83&\underline{26.62}&14.59&\underline{16.05}&\underline{14.21}&29.20&32.31&24.78&14.96&14.51&12.48\\
\midrule
VLN-PETL(ours)&2.81& \textbf{45.96}&51.23&\textbf{42.60}&12.86&\textbf{29.94}&\textbf{27.61}&\textbf{31.81}&37.03&\textbf{27.67}&14.47&\textbf{18.26}&\textbf{15.96}&\underline{30.83}&36.06&\underline{26.73}&14.00&\textbf{15.13}&\textbf{13.03}\\
\bottomrule
\end{tabular}}
\vspace{-1mm}
\caption{Performance of PETL methods on REVERIE. SPL is the main metric for its navigation task, and RGSPL is the main metric for the object grounding task.}
\label{tab:reverie}
\vspace{-7pt}
\end{table*}

%% file: section/conclusion.tex
\section{{Limitations and Future work}}
Though VLN-PETL is proven to be effective on four mainstream VLN tasks of R2R, REVERIE, NDH, and RxR, all these tasks focus on the agent's ability to navigate or ground target object, which has no interactions with the observed objects. Thus, our future work will pay attention to applying PETL methods to other VLN tasks that have interactions with the environment, such as object manipulation.

\section{Conclusion}
In this paper, we present the first study of applying Parameter-Efficient Transfer Learning (PETL) methods to VLN tasks and propose a VLN-specific PETL method named VLN-PETL. Considering the characteristics of VLN, we specifically design two PETL modules to efficiently tune the large pre-trained model for VLN downstream tasks, namely Historical Interaction Booster (HIB) and Cross-modal Interaction Booster (CIB). Both HIB and HIB mainly consist of bottleneck layers and multi-head attention layers, which respectively enhance the vision encoder's learning of history knowledge and the cross-modal encoder's interactions between the language and vision features during the efficient tuning. In addition, we incorporate the vanilla adapters to efficiently tune the language encoder and the LoRA to further improve the integrated performance. Extensive experiments conducted on four mainstream VLN tasks of R2R, REVERIE, NDH, and RxR show the effectiveness of our proposed VLN-PETL. Furthermore, we conduct ablation studies to evaluate the contribution of VLN-PETL components and validate the superiority of our specifically designed HIB and CIB to their counterpart PETL methods.